\documentclass{article}

\usepackage{arxiv}

\usepackage[utf8]{inputenc} 
\usepackage[T1]{fontenc}    
\usepackage{hyperref}       
\usepackage{url}            
\usepackage{booktabs}       
\usepackage{amsfonts}       
\usepackage{nicefrac}       
\usepackage{microtype}      
\usepackage{lipsum}

\usepackage{hyperref}
\usepackage{cleveref}

\usepackage{graphicx}
 \usepackage{algorithm}
\usepackage{algorithmic}

\usepackage{calrsfs}
\DeclareMathAlphabet{\pazocal}{OMS}{zplm}{m}{n}
\newcommand{\La}{\pazocal{L}}
\newcommand{\Ea}{\pazocal{E}}
\newcommand{\Fa}{\pazocal{F}}
\newcommand{\Aa}{\pazocal{A}}
\newcommand{\Ra}{\pazocal{R}}
\newcommand{\Sa}{\pazocal{S}}
\newcommand{\Ga}{\pazocal{G}}
\newcommand{\Xa}{\pazocal{X}}
\newcommand{\Ka}{\pazocal{K}}

\newcommand{\Ca}{\pazocal{C}}
\newcommand{\Pa}{\pazocal{P}}
\newcommand{\Da}{\pazocal{D}}
\newcommand{\Ba}{\pazocal{B}}

\usepackage{amssymb,amsmath}

\title{An Argumentation-based Approach for Explaining Goal Selection in Intelligent Agents}

\author{
  Mariela Morveli-Espinoza \\
  Graduate Program in Electrical and Computer Engineering (CPGEI),\\
 Federal University of Technology - Paran\'{a} (UTFPR),
 Curitiba - Brazil\\
  \texttt{morveli.espinoza@gmail.com} \\
   \And
   Cesar Augusto Tacla \\
  Graduate Program in Electrical and Computer Engineering (CPGEI),\\
 Federal University of Technology - Paran\'{a} (UTFPR),
 Curitiba - Brazil\\
   \texttt{tacla@utfpr.edu.br} \\
   \And
   Henrique Jasisnki \\
   Graduate Program in Electrical and Computer Engineering (CPGEI),\\
 Federal University of Technology - Paran\'{a} (UTFPR),
 Curitiba - Brazil\\
 \texttt{henriquejasinski@alunos.utfpr.edu.br} \\
}

\newtheorem{defn}{Definition}

\newtheorem{exem}{Example}

\begin{document}

\maketitle

\begin{abstract}
During the first step of practical reasoning, i.e. deliberation or goals selection, an intelligent agent generates a set of pursuable goals and then selects which of them he commits to achieve. Explainable Artificial Intelligence (XAI) systems, including intelligent agents, must be able to explain their internal decisions. In the context of goals selection, agents should be able to explain the reasoning path that leads them to select (or not) a certain goal. In this article, we use an argumentation-based approach for generating explanations about that reasoning path. Besides, we aim to enrich the explanations with information about emerging conflicts during the selection process and how such conflicts were resolved.
We propose two types of explanations: the partial one and the complete one and a set of explanatory schemes to generate pseudo-natural explanations. Finally, we apply our proposal to the cleaner world scenario.

\end{abstract}

\keywords{Goal selection  \and Explainable agents \and Formal argumentation}

\section{Introduction}
Practical reasoning means reasoning directed towards actions, i.e. it is the process of figuring out what to do. According to  Wooldridge \hbox{\cite{wooldridge2000reasoning}}, practical reasoning involves two phases: (i) deliberation, which is concerned with deciding what state of affairs an agent wants to achieve, thus, the outputs of deliberation phase are goals the agent intends to pursue, and (ii) means-ends reasoning, which is concerned with deciding how to achieve these states of affairs. The first phase is also decomposed in two parts: (i) firstly, the agent generates a set of pursuable goals\footnote{Pursuable goals are also known as desires and pursued goals as intentions. In this work, we consider that both are goals at different stages of processing, like it was suggested in \cite{castelfranchi2007role}.}, and (ii) secondly, the agent chooses which goals he will be committed to bring about. In this paper, we focus on the first phase, that is, goals selection. 

Given that an intelligent agent may generate multiple pursuable goals, some conflicts among these goals could arise, in the sense that it is not possible to pursue them simultaneously. Thus, a rational agent selects a set of non-conflicting goals based in a criterion or a set of criteria. There are many researches about identifying and resolving such conflict in order to determine the set of pursued goals (e.g., \cite{amgoud2008constrained}\cite{Morveli-Espinoza19}\cite{thangarajah2003detecting}\cite{tinnemeier2007goal}\cite{zatelli2016conflicting}). However, to the best of our knowledge, none of these approaches gives explanations about the reasoning path to determine the final set of pursued goals. Thus, the returned outcomes can be negatively affected due to the lack of clarity and explainability about their dynamics and rationality. 

In order to better understand the problem, consider the well-know ``cleaner world'' scenario, where a set of robots (intelligent agents) has the task of cleaning a dirty environment. The main goal of all the robots is to have the environment clean. Besides cleaning, the robots have other goals such as recharging their batteries or being fixed. Suppose that at a given moment one of the robots (let us call him $\mathtt{BOB}$) detects dirt in slot (5,5); hence, the goal ``cleaning (5,5)'' becomes pursuable. On the other hand, $\mathtt{BOB}$ also auto-detects a technical defect; hence, the goal ``be fixed'' also becomes pursuable. Suppose that $\mathtt{BOB}$ cannot commit to both goals at the same time because the plans adopted for each goal lead to an inconsistency. This means that only one of the goals will become pursued. Suppose that he decides to fix its technical defect instead of cleaning the perceived dirt. During the cleaning task or after the work is finished, the robot can be asked for an explanation about his decision. It is clear that it is important to endow the agents with the ability of explaining their decisions, that is, to explain how and why a certain pursuable goal became (or not) a pursued goal.

Thus, the research questions that are addressed in this paper are: (i) how to endow intelligent agents with the ability of generating explanations about their goals selection process? and (ii) how to improve the informational quality of the explanations?

In addressing the first question, we will use arguments to generate and represent the explanations. At this point, it is important to mention that in this article, argumentation is used in two different ways. Firstly, argumentation will be used in the goals selection process. The input to this process is a set of possible conflicting pursuable goals such that each one has a preference value and a set of plans that allow the agent to achieve them, and the output is a set of pursued goals. We will base on the work of Morveli-Espinoza et al. \cite{Morveli-Espinoza19} for this process. One important contribution given in \cite{Morveli-Espinoza19} is the computational formalization of three forms of conflicts, namely terminal incompatibility, resource incompatibility, and superfluity, which were conceptually defined in \cite{castelfranchi2007role}. The identification of conflicts is done by using plans, which are represented by instrumental arguments\footnote{An instrumental argument is structured like a tree where the nodes are planning rules whose premise is made of a set of sub-goals, resources, actions, and beliefs and its conclusion or claim is a goal, which is the goal achieved by executing the plan represented by the instrumental argument.}. These arguments are compared in order to determine the form of conflict that may exist between them. The set of instrumental arguments and the conflict relation between them make up an Argumentation Framework (AF). Finally, in order to resolve the conflicts, an argumentation semantics is applied. This semantics is a function that takes as input an AF and returns those non-conflicting goals the agent will commit to. Secondly, argumentation is used in the process of explanation generation. The input to this process is the AF mentioned above and the set of pursued goals and the output is a set of arguments that represent explanations. The arguments constructed in this part are not instrumental ones, that is, they do not represent plan but explanations. Regarding the second question, we will use the information in instrumental arguments for enriching explanations about the form(s) of conflict that exists between two goals.

Next section focuses on the knowledge representation and the argumentation process for goal selection. Section \ref{sec-expl} presents the argumentation process for generating explanations. Section \ref{sec-apli} is devoted to the application of the proposal to the cleaner world scenario. Section \ref{relato} presents the main related work. Finally, Section \ref{conclus} is devoted to conclusions and future work.

\section{Argumentation Process for Goals Selection}

In this section, we will present part of the results of the article of Morveli-Espinoza et al. \cite{Morveli-Espinoza19}, on which we will base to construct the explanations. Since we want to enrich the explanations, we will increase the informational capacity of some of the results.

Firstly, let $\La$ be a first-order logical language used to represent the mental states of the agent, $\vdash$ denotes classical inference, and $\equiv$ the logical equivalence. Let $\Ga$ be the set of pursuable goals, which are represented by ground atoms of $\La$ and $\Ba$ be the set of beliefs of the agent, which are represented by ground literals\footnote{Literals are atoms or negation of atoms (the negation of an atom $a$ is denoted $\neg a$).} of $\La$. In order to construct instrumental arguments, other mental states are necessary (e.g. resources, actions, plan rules); however, they are not meaningful in this article. Therefore, we will assume that the knowledge base (denoted by $\Ka$) of an agent includes such mental states, besides his beliefs.

According to Castelfranchi and Paglieri \cite{castelfranchi2007role}, three forms of incompatibility could emerge during the goals selection: terminal, due to resources, and superfluity\footnote{Hereafter, terminal incompatibility is denoted by $t$, resource incompatibility by $r$, and superfluity by $s$.}. Morveli-Espinoza et al. \cite{Morveli-Espinoza19} tackled the problems of identifying and resolving these three forms of incompatibilities. In order to identify these incompatibilities the plans that allow to achieve the goals in $\Ga$ are evaluated. Considering that in their proposal each plan is represented by means of instrumental arguments, as a result of the identification problem, they defined three AFs (one for each form of incompatibility) and a general AF that involves all of the instrumental arguments and attacks of the three forms of incompatibility.

\begin{defn} \textbf{(Argumentation frameworks)} Let $\mathtt{ARG_{ins}}$ be the set of instrumental arguments that an agent can build from $\Ka$\footnote{For further information about how instrumental arguments are built, the reader is referred to \cite{Morveli-Espinoza19}.}. A $x$-AF is a pair $\Aa\Fa_x=\langle \mathtt{ARG}_x, \Ra_x  \rangle$ (for $x \in \{t,r,s\}$) where $\mathtt{ARG}_x \subseteq \mathtt{ARG_{ins}}$ and $\Ra_x$ is the binary relation $\Ra_x$ $\subseteq \mathtt{ARG_{ins}} \times \mathtt{ARG_{ins}}$ that represents the attack between two arguments of $\mathtt{ARG_{ins}}$, so that $(A,B) \in \Ra_x$ denotes that the argument $A$ attacks the argument $B$.
\end{defn}

Since we want to improve the informational quality of explanations, we modify the general AF proposed in \cite{Morveli-Espinoza19} by adding a function that returns the form of incompatibility that exists between two instrumental arguments. Thus, an agent will not only be able to indicate that there is an incompatibility between two goals but he will be able to indicate the form of incompatibility.

\begin{defn} \label{def-gaf}\textbf{(General Argumentation Framework)} Let $\mathtt{ARG_{ins}}$ be a set of instrumental arguments that an agent can build from $\Ka$. A general AF is a tuple $\Aa\Fa_{gen}=\langle \mathtt{ARG_{ins}}, \Ra_{gen}, \mathtt{f\_INCOMP} \rangle$, where $\Ra_{gen}= \Ra_t \cup \Ra_r \cup \Ra_s$ and $\mathtt{f\_INCOMP}: \Ra_{gen} \rightarrow 2^{\{t,r,s\}}$.

\end{defn}

\begin{exem} \label{ejmAFtotal} Recall the cleaner world scenario that was presented in Introduction where agent $\mathtt{BOB}$ has two pursuable goals, which can be expressed as $clean(5,5)$ and $be(fixed)$ in language $\La$. Consider that there are two instrumental arguments whose claim is $clean(5,5)$, namely $A$ that has a sub-argument $E$ whose claim is $pickup(5,5)$ and $C$ that has a sub-argument $D$ whose claim is $mop(5,5)$. Besides, there are two instrumental arguments whose claim is $be(fixed)$, namely $B$ that has a sub-argument $H$ whose claim is $be(in\_workshop)$ and $F$ that does not have any sub-argument.

Recall also that terminal incompatibility was also exemplified. In order to exemplify the other forms of incompatibility and generate the general AF for this scenario, consider the following situations:

\begin{itemize}

\item $\mathtt{BOB}$ has 90 units of battery. He needs 60 units for achieving $C$, he needs 70 units for achieving $A$, he needs 30 units for achieving $B$, and he does not need battery for achieving $F$ because the mechanic can go to his position. We can notice that there is a conflict between $A$ and $B$ and consequently between their sub-arguments.

\item As can be noticed, there are two instrumental arguments whose claim is $clean(5,5)$ and two instrumental arguments whose plan is $be(fixed)$. It would be redundant to perform more than one plan to achieve the same goal, this means that arguments with the same claim are conflicting due to superfluity. This conflict is also extended to their sub-arguments.

\end{itemize}

We can now generate the general AF for the cleaner world scenario: $\Aa\Fa_{gen}=\langle \{A,B, C,D, E,\break F,H\},\Ra_{gen}, \mathtt{f\_INCOMP} \rangle$ where $\Ra_t= \{(A,B), (B,A), (E,B), (B,E), (E,H), (H,E),(A,H),(H,A), (C,B),\break (B,C), (D,B),(B,D), (D,H), (H,D), (C,H),(H,C)\}$, $\Ra_r= \{(A,B), (B,A), (E,B), (B,E), (A,H),(H,A),\break (E,H),(H,E)\}$, and $\Ra_s=\{(C,A), (A,C), (E,D), (D,E), (C,E), (E,C), (A,D), (D,A), (F,B), (B,F), (F,H),\break (H,F)\}$. Figure \ref{aftotal} shows the graph representation.

\begin{figure}[!htb]
	\centering
	\includegraphics[width=0.55\textwidth]{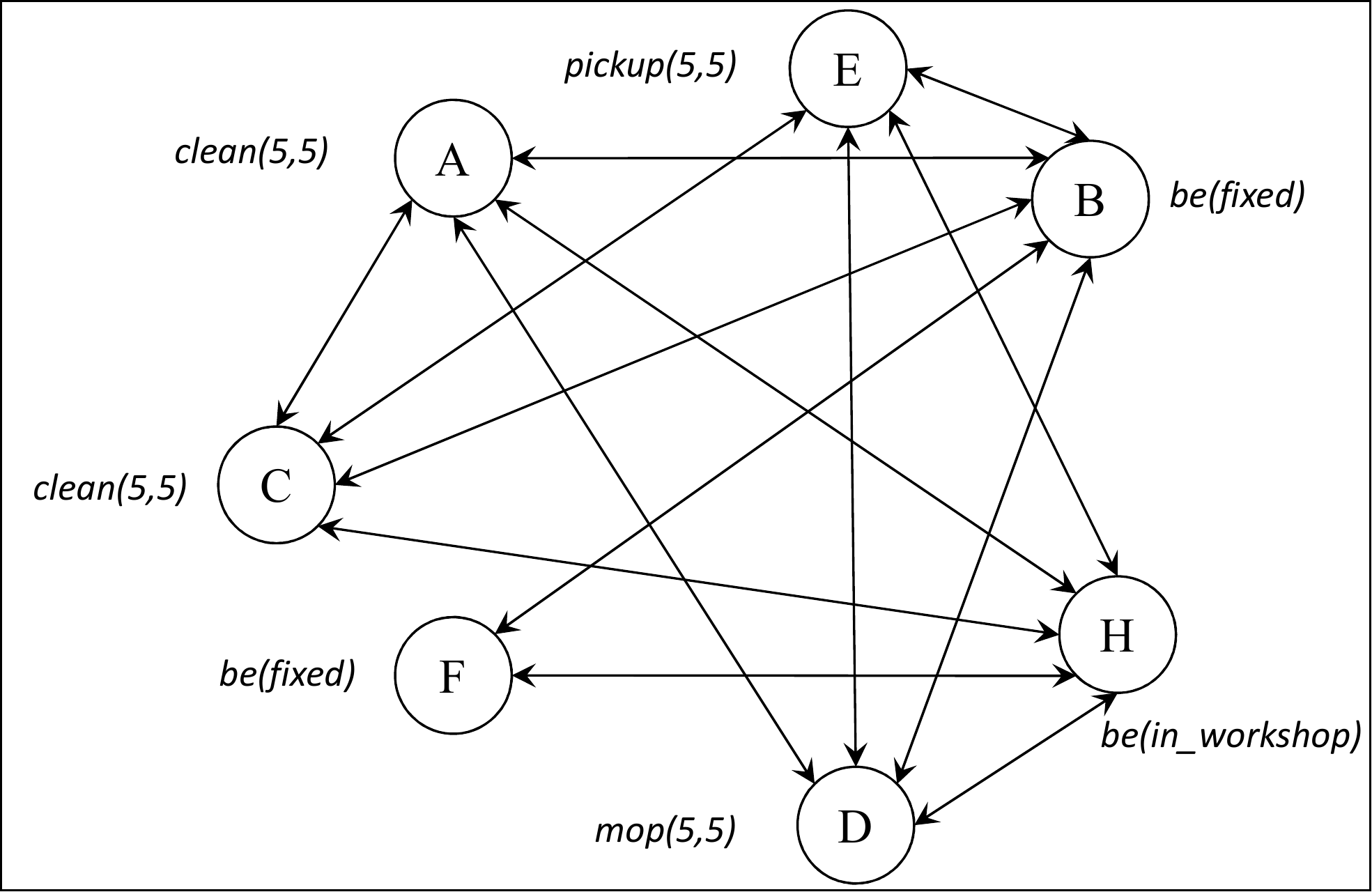} 
	\caption{\textbf{(Obtained from \cite{morveli2019argumentation})} The general AF for the cleaner world scenario. The nodes represent the arguments and the arrows represent the attacks between the arguments. The text next to each node indicates the claim of each instrumental argument.}
	\label{aftotal}
\end{figure}

\end{exem}

So far, we have referred to instrumental arguments -- which represent plans -- however, since the selection is at goals level, it is necessary to generate an AF where arguments represent goals. In order to generate this framework, it is necessary to define when two goals attack each other. This definition is based on the general attack relation $\Ra_{gen}$, which includes the three kinds of attacks that may exist between arguments. Thus, a goal $g$ attacks another goal $g'$ when all the instrumental arguments for $g$ (that is, the plans that allow to achieve $g$) have a general attack relation with all the instrumental arguments for $g'$. This attack relation between goals is captured by the binary relation $\Ra\Ga \subseteq \Ga \times \Ga$. We denote with $(g, g')$ the attack relation between goals $g$ and $g'$. In other words, if $(g,g') \in \Ra\Ga$ means that goal $g$ attacks goal $g'$.

\begin{defn} \label{generalinc}\textbf{(Attack between goals)} Let $\Aa\Fa_{gen}=\langle \mathtt{ARG_{ins}}, \Ra_{gen}, \mathtt{f\_INCOMP} \rangle$ be a general AF, $g, g' \in \Ga$ be two pursuable goals, $\mathtt{ARG\_INS}(g)$\footnote{$\mathtt{ARG\_INS}(g)$ denotes all the instrumental arguments that represent plans that allow to achieve $g$.}$, \mathtt{ARG\_INS}(g') \subseteq \mathtt{ARG_{ins}}$ be the set of arguments for $g$ and $g'$, respectively. Goal $g$ attacks goal $g'$ when $\forall A \in \mathtt{ARG\_INS}(g)$ and $\forall A' \in \mathtt{ARG\_INS}(g')$ it holds that $(A,A') \in \Ra_{gen}$ or $(A',A) \in \Ra_{gen}$. 

\end{defn}

Once the attack relation between two goals was defined, it is also important to determine the forms of incompatibility that exist between any two conflicting goals. The function $\mathtt{INCOMP\_G}(g,g')$ will return the set of forms of incompatibility between goals $g$ and $g'$. Thus, if $(g, g') \in \Ra\Ga$, then $\forall (A,A') \in \Ra_{gen}$ and $\forall (A',A) \in \Ra_{gen}$ where $A \in \mathtt{ARG\_INS}(g)$ and $A' \in \mathtt{ARG\_INS}(g')$, $\mathtt{INCOMP\_G}(g,g') = \bigcup \mathtt{f\_INCOMP}((A,A')) \cup \mathtt{f\_INCOMP}((A',A))$. We can now define an AF where arguments represent goals.

\begin{defn} \textbf{(Goals AF)} \label{def-gaf} An argumentation-like framework for dealing with incompatibility between goals is a tuple $\Ga\Aa\Fa =\langle \Ga, \Ra\Ga, \mathtt{INCOMP\_G}, \mathtt{PREF} \rangle$, where: (i) $\Ga$ is a set of pursuable goals, (ii) $\Ra\Ga \subseteq  \Ga \times \Ga$, (iii) $\mathtt{INCOMP\_G}: \Ra\Ga \rightarrow 2^{\{t,r,s\}}$, and (iv) $\mathtt{PREF}: \Ga \rightarrow (0,1]$ is a function that returns the preference value of a given goal such that 1 stands for the maximum value.

\end{defn}

Hitherto, we have considered that all attacks are symmetrical. However, as can be noticed goals have a preference value, which indicates how valuable each goal is for the agent. Therefore, depending on this preference value, some attacks may be considered successful. This means that the symmetry of the relation attack may be broken.

\begin{defn} \textbf{(Successful attack)}\footnote{In other works (e.g.,  \cite{martinez2006progressive} \cite{modgil2014aspic+}), it is called a defeat relation.} Let $g,g' \in \Ga$ be two goals, we say that $g$ successfully attacks $g'$ when $(g,g') \in \Ra\Ga$ and $\mathtt{PREF}(g) > \mathtt{PREF}(g')$.

Let us denote with $\Ga\Aa\Fa_{sc}=\langle  \Ga, \Ra\Ga_{sc}, \mathtt{INCOMP\_G}, \mathtt{PREF}\rangle$ the AF that results after considering the successful attacks.
\end{defn}

The next step is to determine the set of goals that can be achieved without conflicts, which can also be called acceptable goals and in this article, they can be explicitly called pursued goals. With this aim, it has to be applied an argumentation semantics. Morveli-Espinoza et al. did an analysis about which semantics is more adequate for this problem. They reached to the conclusion that the best semantics is based on conflict-free sets, on which a function is applied. Next we present the definition given in \cite{Morveli-Espinoza19} applied to the Goals AF.

\begin{defn}\label{def-semdel} \textbf{(Semantics)} Given a $\Ga\Aa\Fa_{sc}=\langle  \Ga, \Ra\Ga_{sc}, \mathtt{INCOMP\_G}, \mathtt{PREF}\rangle$. Let $\Sa_{\Ca\Fa}$ be a set of conflict-free sets calculated from $\Ga\Aa\Fa_{sc}$. $\mathtt{MAX\_UTIL}: \Sa_{\Ca\Fa} \rightarrow  2^{\Sa_{\Ca\Fa}}$ determines the set acceptable goals. This function takes as input a set of conflict-free sets and returns those with the maximum utility for the agent in terms of preference value. 

Let $\Ga' \subseteq \Ga$ be the set of goals returned by $\mathtt{MAX\_UTIL}$. This means that $\Ga'$ is the set of goals the agent can commit to, which are called pursued goals or intentions.
\end{defn}

Regarding the function for determining acceptable goals, there may be many ways to make the calculations;
for example, one way of characterizing $\mathtt{MAX\_UTIL}$ is by summing up the preference value of all the goals in an extension. Another way may be by summing up the preference value of just the main goals without considering sub-goals. We will use the first characterization in our scenario.

\begin{exem}\label{ejmfinal} Consider the general AF of Example \ref{ejmAFtotal}, the agent generates: $\Ga\Aa\Fa_{sc}=\langle \{clean(5,5),  pickup(5,5),\break mop(5,5), be(in\_workshop), be(fixed)\}, \{(mop(5,5), pickup(5,5)),$ $(clean(5,5), be(in\_workshop)),  (mop(5,5),\break be(in\_workshop)), (pickup(5,5), be(in\_workshop))\},\mathtt{INCOMP\_G}, \mathtt{PREF}\rangle$. Figure \ref{framegoals} shows this GAF, the preference values of each goal, and the form of incompatibilities that exists between pairs of goals.

\begin{figure}[!h]
	\centering
	\includegraphics[width=0.5\textwidth]{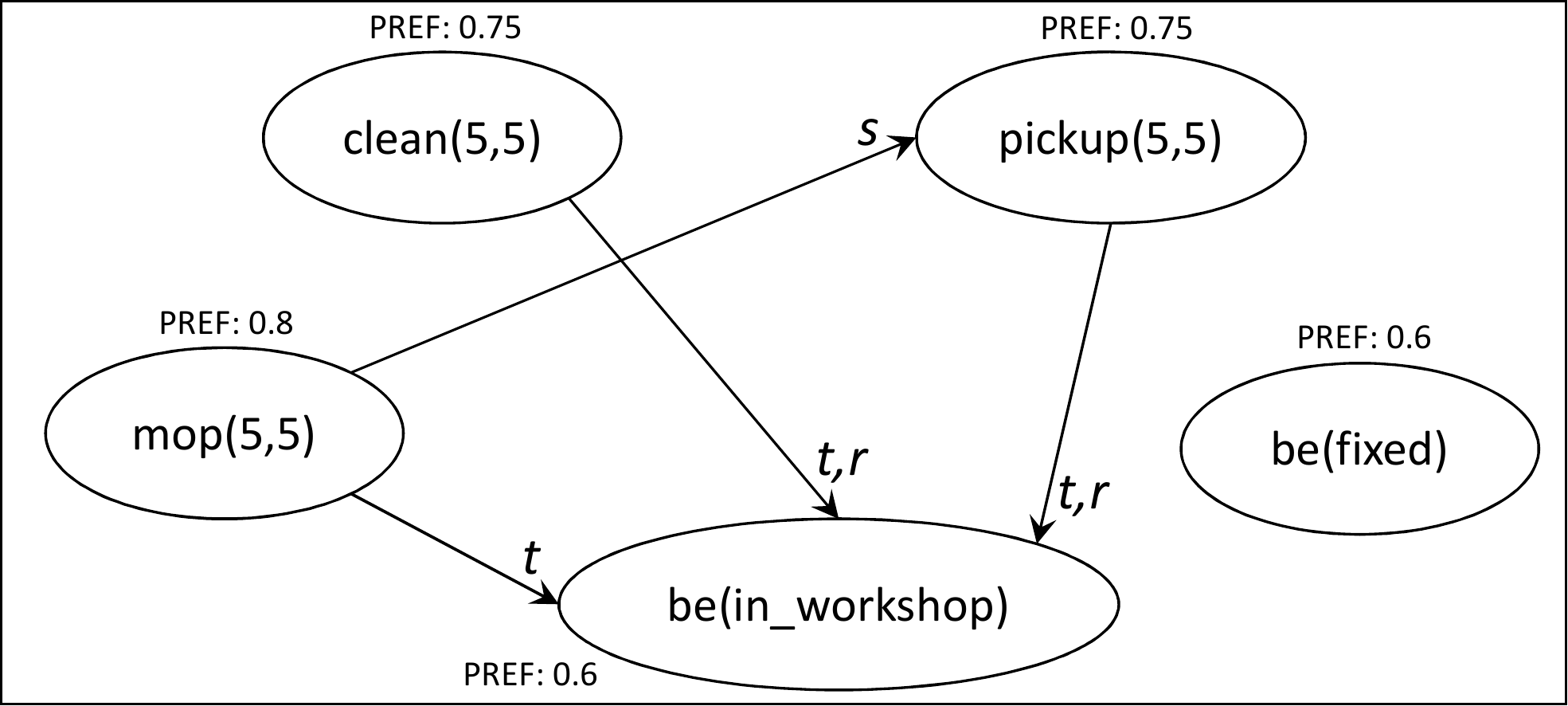} 
	\caption{GAF for the cleaner world scenario. The text next to each arrow indicates the form of incompatibility.}
	\label{framegoals}
\end{figure}

From $\Ga\Aa\Fa_{sc}$, the number of conflict-free extensions is: $|\Sa_{\Ca\Fa}|= 14$. After applying $\mathtt{MAX\_UTIL}$, the extension with the highest preference is: $\{clean(5,5), mop(5,5), be(fixed)\}$. This means that $\Ga'=\{clean(5,5),mop(5,5),be(fixed)\}$ are compatible goals that can be achieved together without conflicts.

\end{exem}

\section{Argumentation Process for Explanations Generation}
\label{sec-expl}
In this section, we present explanatory arguments and the process for generating explanations for a goal become pursued or not.

First of all, let us present the types of questions that can be answered:

\begin{itemize}
\item $\mathtt{WHY}(g)$: it is required an explanation to justify why a goal $g$ became pursued\footnote{In order to better deal with goals, we map each goal to a constant in $\La$.}.
\item $\mathtt{WHY\_NOT}(g)$: it is required an explanation to justify why a goal $g$ did not become pursued.
\end{itemize}

\subsection{Explanatory Arguments and Argumentation Framework}
\label{subsec-delexp}

As a result of the above section, we obtain a Goals Argumentation Framework (GAF) and a set of pursued goals. Recall that in a GAF, the arguments represent goals; hence, in order to generate an explanation from a GAF, it is necessary to generate beliefs and rules -- that reflect the knowledge contained in it -- from which, explanatory arguments can be constructed. Before presenting the beliefs and rules, let us present some functions that will be necessary for the generation of beliefs:

\begin{itemize}

\item $\mathtt{COMPS}(\Ga\Aa\Fa_{sc})=\{g \mid  \nexists (g, g') \in \Ra\Ga_{sc}$ (or $(g',g) \in \Ra\Ga_{sc})$, where $g, g' \in \Ga \}$ . This function returns the set of goals without conflicting relations.

\item $\mathtt{EVAL\_PREF}(\Ga\Aa\Fa_{sc})=\{(g,g') \mid (g,g') \in \Ra\Ga_{sc}$ and $ (g',g) \not\in \Ra\Ga_{sc}\}$. This function returns all the pairs of goals in $\Ra\Ga'$ that represent non-symmetrical relations between goals. When the relation is not symmetrical, it means that one of the goals is preferred to the other.


\end{itemize}

Using these functions, the set of beliefs generated from a $\Ga\Aa\Fa_{sc} =\langle \Ga, \Ra\Ga_{sc}, \mathtt{INCOMP\_G}, \mathtt{PREF} \rangle$ are the following:

\begin{itemize}
\item $\forall g \in \mathtt{COMPS}(\Ga\Aa\Fa_{sc})$ generate a belief $\neg incomp(g)$
\item $\forall(g,g') \in \mathtt{EVAL\_PREF}(\Ga\Aa\Fa_{sc})$, if $\mathtt{PREF}(g) > \mathtt{PREF}(g')$, then generate $pref(g,g')$ and $\neg pref(g',g)$. 
\item $\forall(g,g') \in (\Ra\Ga_{sc} \setminus \mathtt{EVAL\_PREF}(\Ga\Aa\Fa_{sc}))$ generate a belief $eq\_pref(g,g')$. These beliefs are created for those pairs of goals with equal preference. 
\item $\forall (g,g') \in \Ra\Ga_{sc}$ generate a belief $incompat(g,g',ls)$ where $ls=\mathtt{INCOMP\_G}(g,g')$
\item $\forall g \in \Ga'$ generate a belief $max\_util(g)$
\item $\forall g \in \Ga \setminus \Ga'$ generate a belief $\neg max\_util(g)$
\end{itemize}

All the beliefs that are generated have to be added to the set of beliefs $\Ba$ of the agent. These beliefs are necessary for triggering any of the following rules:

\vspace*{-0.3cm}

\begin{itemize}
\item $r1: \neg incomp(x) \rightarrow pursued(x)$
\item $r2: incompat(x, y,ls) \wedge pref(x,y) \rightarrow pursued(x)$
\item $r3: incompat(x, y,ls) \wedge \neg pref(y,x) \rightarrow \neg pursued(y)$
\item $r4: incompat(x, y,ls) \wedge eq\_pref(x,y) \rightarrow pursued(x)$
\item $r5: max\_util(x) \rightarrow pursued(x)$
\item $r6: \neg max\_util(x) \rightarrow \neg pursued(x)$
\end{itemize}

\vspace*{-0.25cm}

Let $\Ea\Ra =\{r1, r2,r3,r4,r5,r6\}$ be the set of rules necessary for constructing explanatory arguments.

\begin{defn}\textbf{(Explanatory argument)} Let $\Ba$, $\Ea\Ra$, and $g \in \Ga$ be the set of beliefs, set of rules, and a goal of an agent, respectively. An explanatory argument constructed from $\Ba$ and $\Ea\Ra$ for determining the status of $g$ is a pair $A=\langle \Sa, h \rangle$ such that (i) $\Sa \subseteq \Ba \cup \Ea\Ra$, (ii) $h \in \{pursued(g), \neg pursued(g)\}$, (iii) $\Sa \vdash h$, and (iv) $\Sa$ is consistent and minimal for the set inclusion\footnote{Minimal means that there is no $\Sa' \subset \Sa$ such that $\Sa \vdash h$ and consistent means that it is not the case that $\Sa \vdash pursued(g)$ and $\Sa \vdash \neg pursued(g)$ \cite{hunter2010base}.}.

\noindent Let $\mathtt{ARG}_{exp}$ be the set of explanatory arguments that can be built from $\Ba$ and $\Ea\Ra$. We call $\Sa$ the support of an argument $A$ (denoted by $\mathtt{SUPPORT}(A)$) and $h$ its claim (denoted by $\mathtt{CLAIM}(A)$).
\end{defn}

We can notice that rules in $\Ea\Ra$ can generate conflicting arguments because they have inconsistent conclusions. Thus, we need to define the concept of attack. In this context, the attack that can exist between two explanatory arguments is the well-known rebuttal \cite{besnard2009argumentation}, where two explanatory arguments support contradictory claims. Formally:

\begin{defn} \textbf{(Rebuttal)} Let $\langle\Sa, h \rangle$ and $\langle \Sa', h' \rangle$ be two explanatory arguments. $\langle\Sa, h \rangle$ rebuts $\langle\Sa', h' \rangle$ iff $h \equiv \neg h'$.

\end{defn}

Rebuttal attack has a symmetric nature, this means that two arguments rebut each other, that is, they mutually attack. Recall that the semantics for determining the set of pursued goals is based on conflict-free sets and on a function based on the preference value of the goals. This function is decisive in the selection of the extension that includes the goals the agent can commit to. Thus, it is natural to believe that arguments related to such function are stronger than other arguments. This difference in the strength of arguments turns out in a defeat relation between them, which breaks the previously mentioned symmetry. 

\begin{defn} \textbf{(Defeat Relation - $\Da$)} Let $\Ea\Ra$ be the set of rules and $A=\langle\Sa, h \rangle$ and $B=\langle \Sa', h' \rangle$ be two explanatory arguments such that $A$ rebuts $B$ and vice versa. $A$ defeats $B$ iff $r5 \in \Sa$ (or $r6 \in \Sa$). 

We denote with $(A,B)$ the defeat relation between $A$ and $B$. In other words, if $(A,B) \in \Da$, it means that $A$ defeats $B$.

\end{defn}

Once we have defined arguments and the defeat relation, we can generate the AF. It is important to make it clear that a different AF is generated for each goal.

\begin{defn} \textbf{(Explanatory Argumentation Framework)} Let $g \in \Ga$ be a pursuable goal. An Explanatory AF for $g$ is a pair $\Xa\Aa\Fa_g=\langle \mathtt{ARG}_{exp}^g, \Da^g \rangle$ where:

\begin{itemize}

\item $\mathtt{ARG}_{exp}^g \subseteq \mathtt{ARG}_{exp}$ such that $\forall A \in \mathtt{ARG}_{exp}^g$, $\mathtt{CLAIM}(A)=pursued(g)$ or $\mathtt{CLAIM}(A)=\neg pursued(g)$.

\item $\Da^g \subseteq \mathtt{ARG}_{exp}^g \times \mathtt{ARG}_{exp}^g$ is a binary relation that captures the defeat relation between arguments in $\mathtt{ARG}_{exp}^g$.

\end{itemize}

\end{defn}

The next step is to evaluate the arguments that make part of the AF. This evaluation is important because it determines the set of non-conflicting arguments, which in turn determines if a goal becomes pursued or not. Recall that for obtaining such set, an argumentation semantics has to be applied. Unlike the semantics for goals selection, in this case we can use any of the semantics defined in literature. Next, the main semantics introduced by Dung \cite{dung1995acceptability} are recalled\footnote{It is not the scope of this article to study the most adequate semantics for this context or the way to select an extension when more than one is returned by a semantics. }.

\begin{defn} \textbf{(Semantics)} Let $\Xa\Aa\Fa_g=\langle \mathtt{ARG}_{exp}^g, \Da^g \rangle$ be an explanatory AF and $\Ea \subseteq \mathtt{ARG}_{exp}^g$:

\begin{itemize}

\item $\Ea$ is \textbf{\textit{conflict-free}} if $\forall A,B \in \Ea$, $(A,B) \notin \Da^g$ 
\item $\Ea$ \textbf{\textit{defends}} $A$ iff $\forall B \in \mathtt{ARG}_{exp}^g$, if $(B, A) \in \Da^g$, then $ \exists C \in \Ea$ s.t. $(C,B) \in \Da^g$.
\item $\Ea$ is \textbf{\textit{admissible}} iff it is conflict-free and defends all its elements. 
\item A conflict-free $\Ea$ is a \textbf{\textit{complete extension}} iff we have $\Ea=\{A | \Ea$ defends $A\}$. 
\item $\Ea$ is a \textbf{\textit{preferred extension}} iff it is a maximal (w.r.t. the set inclusion) complete extension.
\item $\Ea$ is a \textbf{\textit{grounded extension}} iff is a minimal (w.r.t. set inclusion) complete extension.
\item $\Ea$ is a \textbf{\textit{stable extension}} iff $\Ea$ is conflict-free and $\forall A \in$ $\mathtt{ARG}_{exp}^g$, $\exists B \in \Ea$ such that $(B,A) \in \Da^g$.

\end{itemize}
\end{defn}

Finally, a goal $g$ becomes pursued when $\exists A \in \Ea$ such that $\mathtt{CLAIM}(A)=pursued(g)$.

\subsection{Explanation Generation Process}

In this article, an explanation is made up of a set of explanatory arguments that justify the fact that a pursuable goal becomes (or not) pursued. Recall that there is a different explanatory AF for each pursuable goal. Thus, we can say that an explanation for a given goal $g$ is given by the explanatory AF generated for it, that is $\Xa\Aa\Fa_g$. Besides, if $g \in \Ga'$, the explanation is required by using $\mathtt{WHY}(g)$; otherwise, the explanation is required by using $\mathtt{WHY\_NOT}(g)$.
Finally, we can differentiate between partial and complete explanations depending on the set of explanatory arguments that are employed for the justification:

\begin{itemize}
\item A \textit{complete explanation} for $g$ is: $\Ca\Ea_g=\Xa\Aa\Fa_g$
\item A \textit{partial explanation} for $g$ is: $\Pa\Ea_g=\Ea$, where $\Ea$ is an extension obtained by applying a semantics to $\Xa\Aa\Fa_g$.
\end{itemize}

We can now present the steps for generating explanations. Given a $\Ga\Aa\Fa_{sc} =\langle \Ga, \Ra\Ga_{sc}, \mathtt{INCOMP\_G}, \mathtt{PREF} \rangle$ and a set of pursued goals $\Ga'$, the steps for generating an explanation for a goal $g \in \Ga$ are:

\begin{enumerate}
\item From $\Ga\Aa\Fa_{sc}$ generate the respective beliefs and add to $\Ba$
\item Trigger the rules in $\Ea\Ra$ that can be unified with the beliefs of $\Ba$ 
\item Construct explanatory arguments based on the rules and beliefs of the two previous items
\item $\forall g \in \Ga$ do 
\begin{enumerate}
\item Generate the respective explanatory AF (that is, $\Xa\Aa\Fa_g$) with the arguments whose claim is $pursued(g)$ or $\neg pursued(g)$ and the defeat relation
\item Calculate the extension $\Ea$ from $\Xa\Aa\Fa_g$
\end{enumerate}

\end{enumerate}

\subsection{From Explanatory Arguments to Explanatory Sentences}

Like it was done in \cite{guerrero2016activity}, in this sub-section we present a pseudo-natural language for improving the understanding of the explanations when the agents are interacting with human users.  Thus, we propose a set of \textit{explanatory schemes}, one for each rule in $\Ea\Ra$. This means that depending on which rule an argument was constructed, the explanation scheme is different. In this first version of the scheme, we will generate explanatory sentences only for partial explanations.  

Recall that goals are mapped to constants of $\La$, in order to improve the natural language let $\mathtt{NAME}(g)$ denote the original predicate of a given goal $g$. Besides, let $\mathtt{RULE}(A)$ denote which of the rules in $\Ea\Ra$ was employed in order to construct $A$. 

\begin{defn} \textbf{(Explanatory Schemes)} Let $A=\langle \Sa, h \rangle$ be an explanatory argument. An explanatory scheme $\mathtt{exp\_sch}$ for $A$ is:\footnote{Underlined characters represent the variables of the schemes, which depend on the variables of rules.}

\begin{itemize}
\item If $\mathtt{RULE}(A)=r1:\neg incomp(x) \rightarrow pursued(x)$, then\\
 $\mathtt{exp\_sch}= \langle\mathtt{\underline{NAME}}(\underline{x})$ \textit{has no incompatibility, so it became pursued.}$\rangle$ 
\item If $\mathtt{RULE}(A)=r2:incompat(x, y,ls) \wedge pref(x,y) \rightarrow pursued(x)$, then\\ $\mathtt{exp\_sch}= \langle \mathtt{\underline{NAME}}(\underline{x})$ and $\mathtt{\underline{NAME}}(\underline{y})$ have the following conflicts: $\underline{ls}$. Since $\mathtt{\underline{NAME}}(\underline{x})$ is more preferable than $\mathtt{\underline{NAME}}(\underline{y})$, $\mathtt{\underline{NAME}}(\underline{x})$ became pursued.$\rangle$
\item If $\mathtt{RULE}(A)=r3: incompat(x, y,ls) \wedge \neg pref(y,x) \rightarrow \neg pursued(y)$, then\\
$\mathtt{exp\_sch}= \langle \mathtt{\underline{NAME}}(\underline{x})$ and $\mathtt{\underline{NAME}}(\underline{y})$ have the following conflicts: $\underline{ls}$. Since $\mathtt{\underline{NAME}}(\underline{y})$ is less preferable than $\mathtt{\underline{NAME}}(\underline{x})$, $\mathtt{\underline{NAME}}(\underline{y})$ did not become pursued.$\rangle$
\item If $\mathtt{RULE}(A)=r4: incompat(x, y,ls) \wedge eq\_pref(x,y) \rightarrow pursued(x)$, then\\
$\mathtt{exp\_sch}= \langle \mathtt{\underline{NAME}}(\underline{x})$ and $\mathtt{\underline{NAME}}(\underline{y})$ have the following conflicts: $\underline{ls}$. Since $\mathtt{\underline{NAME}}(\underline{x})$ and $\mathtt{\underline{NAME}}(\underline{y})$ have the same preference value, $\mathtt{\underline{NAME}}(\underline{x})$ became pursued.$\rangle$
\item If $\mathtt{RULE}(A)=r5: max\_util(x) \rightarrow pursued(x)$, then\\
$\mathtt{exp\_sch}= \langle$Since $\mathtt{\underline{NAME}}(\underline{x})$ belonged to the set of goals that maximize the utility, it became pursued.$\rangle$
\item If $\mathtt{RULE}(A)=r6: \neg max\_util(x) \rightarrow \neg pursued(x)$, then\\
$\mathtt{exp\_sch}= \langle$Since $\mathtt{\underline{NAME}}(\underline{x})$ did not belong to the set of goals that maximizes the utility, it did not become pursued.$\rangle$

\end{itemize}

\end{defn}

\section{Application: Cleaner World Scenario}
\label{sec-apli}
Let us consider the $\Ga\Aa\Fa_{sc} =\langle \Ga, \Ra\Ga_{sc}, \mathtt{INCOMP\_G}, \mathtt{PREF} \rangle$ presented in Example 2, whose graph is depicted in Figure \ref{framegoals}. Recall also that $\Ga'=\{clean(5,5),mop(5,5), be(fixed)\}$. 

Firstly, we map the goals in $\Ga$ into constants of $\La$ in the following manner: $g_1=clean(5,5)$, $g_2=pickup(5,5)$, $g_3=mop(5,5)$, $g_4=be(in\_workshop)$, and $g_5=be(fixed)$. We will also map the beliefs and rules to constants in $\La$.

We can now follow the steps to generate the explanations:
\vspace*{0.2cm}

\noindent\textbf{1. Generate beliefs}

- $b_1:\neg incomp(g_5)$ \hspace*{3.1cm} $b_{10}: \neg max\_util(g_4)$\\
\indent- $b_2: incompat(g_3,g_2, `s$'$)$ \hspace*{2cm} $b_{11}: pref(g_3, g_4)$\\
\indent- $b_3: incompat(g_3,g_4, `t$'$)$ \hspace*{2.05cm} $b_{12}: \neg pref(g_4, g_3)$\\
\indent- $b_4: incompat(g_1,g_4, `t,r$'$)$ \hspace*{1.75cm} $b_{13}: pref(g_1, g_4)$\\
\indent- $b_5: incompat(g_2,g_4, `t,r$'$)$ \hspace*{1.75cm} $b_{14}: \neg pref(g_4, g_1)$\\
\indent- $b_6: max\_util(g_1)$ \hspace*{3.15cm} $b_{15}: pref(g_2, g_4)$\\
\indent- $b_7: max\_util(g_3)$ \hspace*{3.15cm} $b_{16}: \neg pref(g_4, g_2)$\\
\indent- $b_8: max\_util(g_5)$ \hspace*{3.15cm} $b_{17}: pref(g_3, g_2)$\\
\indent- $b_9: \neg max\_util(g_2)$ \hspace*{2.9cm} $b_{18}: \neg pref(g_2, g_3)$

\vspace*{0.2cm}

\noindent\textbf{2. Trigger rules}

- $r_1: \neg incomp(g_5) \rightarrow pursued(g_5)$\\
\indent - $r_2: incompat(g_3,g_2, `s$'$) \wedge pref(g_3,g_2) \rightarrow pursued(g_3)$\\
\indent - $r_3: incompat(g_3,g_2, `s$'$) \wedge \neg pref(g_2,g_3) \rightarrow \neg pursued(g_2)$\\
\indent - $r_4: incompat(g_3,g_4, `t$'$) \wedge pref(g_3,g_4) \rightarrow pursued(g_3)$\\
\indent - $r_5: incompat(g_3,g_4, `t$'$) \wedge \neg pref(g_4,g_3) \rightarrow \neg pursued(g_4)$\\
\indent - $r_6: incompat(g_1,g_4, `t,r$'$) \wedge pref(g_1,g_4) \rightarrow pursued(g_1)$\\
\indent - $r_7: incompat(g_1,g_4, `t,r$'$) \wedge \neg pref(g_4,g_1) \rightarrow \neg pursued(g_4)$\\
\indent - $r_8: incompat(g_2,g_4, `t,r$'$) \wedge pref(g_2,g_4) \rightarrow pursued(g_2)$\\
\indent - $r_9: incompat(g_2,g_4, `t,r$'$) \wedge \neg pref(g_4,g_2) \rightarrow \neg pursued(g_4)$\\
\indent - $r_{10}:max\_util(g_1) \rightarrow pursued(g_1)$\\
\indent - $r_{11}:max\_util(g_3) \rightarrow pursued(g_3)$ \\
\indent - $r_{12}:max\_util(g_5) \rightarrow pursued(g_5)$\\
\indent - $r_{13}:\neg max\_util(g_2) \rightarrow \neg pursued(g_2)$ \\
\indent - $r_{14}:\neg max\_util(g_4) \rightarrow \neg pursued(g_4)$

\vspace*{0.2cm}

\noindent\textbf{3. Construct explanatory arguments}

- $A_1=\langle \{b_1, r_1\}, pursued(g_5)\} \rangle$ \hspace*{1cm} - $A_2=\langle \{b_2, b_{17}, r_2\}, pursued(g_3)\} \rangle$\\
\indent - $A_3=\langle \{b_2, b_{18}, r_3\}, \neg pursued(g_2)\} \rangle$ \hspace*{0.18cm} - $A_4=\langle \{b_3, b_{11}, r_4\}, pursued(g_3)\} \rangle$\\
\indent - $A_5=\langle \{b_3, b_{12}, r_5\}, \neg pursued(g_4)\} \rangle$ \hspace*{0.18cm} - $A_6=\langle \{b_4, b_{13}, r_6\}, pursued(g_1)\} \rangle$\\
\indent - $A_7=\langle \{b_4, b_{14}, r_7\}, \neg pursued(g_4)\} \rangle$ \hspace*{0.18cm} - $A_8=\langle \{b_5, b_{15}, r_8\}, pursued(g_2)\} \rangle$\\
\indent - $A_9=\langle \{b_5, b_{16}, r_9\}, \neg pursued(g_4)\} \rangle$ \hspace*{0.18cm} - $A_{10}=\langle \{b_6, r_{10}\}, pursued(g_1)\} \rangle$\\
\indent - $A_{11}=\langle \{b_7, r_{11}\}, pursued(g_3)\}$ $\rangle$ \hspace*{0.75cm }- $A_{12}=\langle \{b_8, r_{12}\}, pursued(g_5)\} \rangle$\\
\indent - $A_{13}=\langle \{b_9, r_{13}\}, \neg pursued(g_2)\} \rangle$\hspace*{0.75cm }- $A_{14}=\langle \{b_{10}, r_{14}\}, \neg pursued(g_4)\} \rangle$

\vspace*{0.2cm}

\noindent\textbf{4. For each goal, generate an explanatory AF and extension }

- For $g_1$: $\Xa\Aa\Fa_{g_1}= \langle \{A_6, A_{10}\}, \{\} \rangle$, $\Ea=\{A_6, A_{10}\}$\\
\indent - For $g_2$: $\Xa\Aa\Fa_{g_2}= \langle \{A_3, A_8, A_{13}\}, \{(A_3,A_8), (A_{13},A_8)\} \rangle$, $\Ea=\{A_3, A_{13}\}$\\
\indent - For $g_3$: $\Xa\Aa\Fa_{g_3}= \langle \{A_2,A_4, A_{11}\}, \{\} \rangle$, $\Ea=\{A_2,A_4, A_{11}\}$\\
\indent - For $g_4$: $\Xa\Aa\Fa_{g_4}= \langle \{A_5, A_7, A_9, A_{14}\}, \{\} \rangle$, $\Ea=\{A_5, A_7, A_9, A_{14}\}$\\
\indent - For $g_5$: $\Xa\Aa\Fa_{g_5}= \langle \{A_1, A_{12}\}, \{\} \rangle$, $\Ea=\{A_1, A_{12}\}$\\

Thus, the -- partial or complete -- explanations for justifying the status of each goal were generated. Next, we present the query, set of arguments of the partial explanation, and the explanatory sentences for the status of each goal:

\begin{itemize}
\item For the query $\mathtt{WHY}(g_1)$, we have $\Pa\Ea=\{A_6, A_{10}\}$, which can be written:\\
 * $\underline{clean(5,5)}$ \textit{and} $\underline{be(in\_workshop)}$\textit{ have the following conflicts:} \underline{$`t,r$'}. \textit{Since} $\underline{clean(5,5)}$ \textit{is more preferable than} $\underline{be(in\_workshop)}$, $\underline{clean(5,5)}$ \textit{became pursued}\\
 * \textit{Since} $\underline{clean(5,5)}$ \textit{belonged to the set of goals that maximizes the utility,} it \textit{became pursued}

 \item For the query $\mathtt{WHY\_NOT}(g_2)$, we have $\Pa\Ea=\{A_3, A_{13}\}$, which can be written:\\
 * $\underline{mop(5,5)}$ \textit{and} $\underline{pickup(5,5)}$ \textit{have the following conflicts: }\underline{$`s$'}. \textit{Since} $\underline{pickup(5,5)}$ \textit{is less preferable than} $\underline{mop(5,5)}$, $\underline{pickup(5,5)}$ \textit{did not become pursued}\\
 * \textit{Since} $\underline{pickup(5,5)}$ \textit{did not belong to the set of goals that maximizes the utility, it did not become pursued}
 
 \item For the query $\mathtt{WHY}(g_3)$, we have $\Pa\Ea=\{A_2,A_4, A_{11}\}$, which can be written:\\
 * $\underline{mop(5,5)}$ \textit{and} $\underline{pickup(5,5)}$\textit{ have the following conflicts:} \underline{$`s$'}. \textit{Since} $\underline{mop(5,5)}$ \textit{is more preferable than} $\underline{pickup(5,5)}$, $\underline{mop(5,5)}$ \textit{became pursued}\\
 * $\underline{mop(5,5)}$ \textit{and} $\underline{be(in\_workshop)}$ \textit{have the following conflicts:} \underline{$`t$'}. \textit{Since} $\underline{mop(5,5)}$ \textit{is more preferable than} $\underline{be(in\_workshop)}$, $\underline{mop(5,5)}$ \textit{became pursued}\\
 * \textit{Since} $\underline{mop(5,5)}$ \textit{belonged to the set of goals that maximizes the utility,} it \textit{became pursued}

\item For the query $\mathtt{WHY\_NOT}(g_4)$, we have $\Pa\Ea=\{A_5, A_7, A_9, A_{14}\}$, which can be written:\\
* $\underline{mop(5,5)}$ \textit{and} $\underline{be(in\_workshop)}$ \textit{have the following conflicts: }\underline{$`t$'}. \textit{Since} $\underline{be(in\_workshop)}$ \textit{is less preferable than} $\underline{mop(5,5)}$, $\underline{be(in\_workshop)}$ \textit{did not become pursued}\\
* $\underline{clean(5,5)}$ \textit{and} $\underline{be(in\_workshop)}$ \textit{have the following conflicts: }\underline{$`t,r$'}. \textit{Since} $\underline{be(in\_workshop)}$ \textit{is less preferable than} $\underline{clean(5,5)}$, $\underline{be(in\_workshop)}$ \textit{did not become pursued}\\
* $\underline{pickup(5,5)}$ \textit{and} $\underline{be(in\_workshop)}$ \textit{have the following conflicts: }\underline{$`t,r$'}. \textit{Since} $\underline{be(in\_workshop)}$ \textit{is less preferable than} $\underline{pickup(5,5)}$, $\underline{be(in\_workshop)}$ \textit{did not become pursued}\\
 * \textit{Since} $\underline{be(in\_workshop)}$ \textit{did not belong to the set of goals that maximizes the utility, it did not become pursued}

\item For the query $\mathtt{WHY}(g_5)$, we have $\Pa\Ea=\{A_1, A_{12}\}$, which can be written:\\
 * $\underline{be\_fixed}$ \textit{has no incompatibility, so it became pursued}\\
 * \textit{Since} $\underline{be\_fixed}$ \textit{belonged to the set of goals that maximizes the utility,} it \textit{became pursued}

\end{itemize}

For all the queries, except $\mathtt{WHY\_NOT}(g_2)$, the complete explanation is the same. In the case of  $\mathtt{WHY\_NOT}(g_2)$, the complete explanation includes the attack relations between some of the arguments of its explanatory AF.

\begin{figure}[!htb]
	\centering
	\includegraphics[width=1\textwidth]{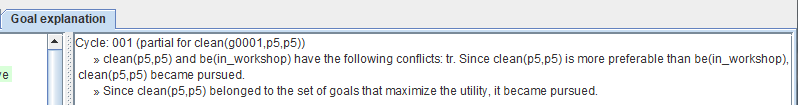} 
	\caption{Partial explanation for query $\mathtt{WHY}(g_1)$. Obtained by using the simulator ArgAgent.}
	\label{parcial}
\end{figure}

We are also working in a simulator -- called ArgAgent\footnote{Available at: https://github.com/henriquermonteiro/BBGP-Agent-Simulator/} -- for generating explanations. In it first version, just partial explanations are generated. Figure \ref{parcial} shows the explanation for query $\mathtt{WHY}(g_1)$.

\section{Related Work}
\label{relato}
Since XAI is a recently emerged domain in Artificial Intelligence, there are few reviews about the works in this area. In \cite{anjomshoae2019explainable}, Anjomshoae et al. make a Systematic Literature Review about goal-driven XAI, i.e., explainable agency for robots and agents. Their results show that 22\% of the platforms and architectures have not explicitly indicate their method for generating explanations, 18\% of papers relied on \textit{ad-hoc} methods, 9\% implemented their explanations in BDI architecture. 

Some works relied on the BDI model are the following. In \cite{broekens2010you} and \cite{harbers2010design}, Broekens et al. and Harbers et al., respectively, focus on generating explanations for humans about how their goals were achieved. Unlike our proposal, their explanations do not focus on the goals selection. Langley et al. \cite{langley2017explainable} focus on settings in which an agent receives instructions, performs them, and then describes and explains its decisions and actions afterwards. 

Sassoon et al. \cite{sassoonexplainable} propose an approach of explainable argumentation based on argumentation schemes and argumentation-based dialogues. In this approach, an agent provides explanations to patients (human users) about their treatments. In this case, argumentation is applied in a different way than in our proposal and with other focus, they generate explanations for information seeking and persuasion. Finally, Morveli-Espinoza et al. \cite{morveliespinoza2019argumentation} propose an argumentation-based approach for generating explanations about the intention formation process, that is, since a goal is a desire until it becomes an intention; however, the generated explanations about goals selection are not detailed and they do not present a pseudo-natural language.

\section{Conclusions and Future Work}
\label{conclus}

In this article, we presented an argumentation-based approach for generating explanations about the goals selection process, that is, giving reasons to justify the transition of a set of goals from being pursuable (desires) to pursued (intentions). Such reasons are related to the conflicts that may exist between pursuable goals and how that conflicts were resolved. In the first part of the approach, argumentation was employed to deal with conflicts and in the second part it was employed to generate explanations. In order to improve the informational quality of explanations, we extended the results presented in \cite{espinoza2019argumentation}. Thus, explanations also include the form of incompatibility that exists between goals. Besides, we proposed a pseudo-natural language that is a first step to generate explanations for human users. Therefore, our proposal is able generate explanations for both intelligent agents and human-users.

As future work, we aim to further improve the informational quality of explanations by allowing information seeking about the exact point of conflict between two instrumental arguments (or plans) and information about the force of the arguments. The pseudo-natural language was only applied to partial explanations, we plan to extend such language in order to support complete explanations.

\section*{Acknowledgment}
This work is partially founded by CAPES (Coordena\c{c}\~{a}o de Aperfei\c{c}oamento de Pessoal de N\'{i}vel Superior).

\bibliographystyle{unsrt}  
\bibliography{IEEEtran}  


\end{document}